\documentclass{ifacconf}  

\usepackage{amsmath}
\usepackage{bm}
\usepackage{color}
\usepackage{multirow}
\usepackage{graphicx}
\usepackage[round,numbers]{natbib}
\usepackage{amssymb}
\DeclareMathOperator*{\argmin}{argmin}

\begin{document}
\begin{frontmatter}

\title{Cooperative Localisation of a GPS-Denied UAV in 3-Dimensional Space Using Direction of Arrival Measurements\thanksref{footnoteinfo}
}

\thanks[footnoteinfo]{This work was supported by the Australian Research Council (ARC) under the ARC grants \mbox{DP-130103610} and \mbox{DP-160104500}, and by Data61-CSIRO (formerly NICTA).}

\author[First]{James Russell} 
\author[First]{Mengbin Ye} 
\author[First,Second,Third]{Brian D. O. Anderson}
\author[Fourth]{Hatem Hmam}
\author[Fourth]{Peter Sarunic}

\address[First]{Research School of Engineering, Australian National University, Canberra, A.C.T. 2601, Australia}
\address[Second]{Hangzhou Dianzi University, Hangzhou, Zhejiang, China}
\address[Third]{Data61-CSIRO (formerly NICTA Ltd.) }
\address[Fourth]{Australian Defence Science and Technology Group (DST Group)}
\address{E-mail: \{u5542624, mengbin.ye, brian.anderson\}@anu.edu.au} 
\address{\{hatem.hmam, peter.sarunic\}@dsto.defence.gov.au}

\begin{abstract}
This paper presents a novel approach for localising a GPS (Global Positioning System)-denied Unmanned Aerial Vehicle (UAV) with the aid of a GPS-equipped UAV in three-dimensional space. The GPS-equipped UAV makes discrete-time broadcasts of its global coordinates. The GPS-denied UAV simultaneously receives the broadcast and takes direction of arrival (DOA) measurements towards the origin of the broadcast in its local coordinate frame (obtained via an inertial navigation system (INS)). The aim is to determine the difference between the local and global frames, described by a rotation and a translation. In the noiseless case, global coordinates were recovered exactly by solving a system of linear equations. When DOA measurements are contaminated with noise, rank relaxed semidefinite programming (SDP) and the Orthogonal Procrustes algorithm are employed. Simulations are provided and factors affecting accuracy, such as noise levels and number of measurements, are explored. 
\end{abstract}


\end{frontmatter}


\section{Introduction}

Unmanned aerial vehicles play a central role in many defence reconnaissance and surveillance operations. Formations of UAVs can provide greater reliability and coverage when compared to a single UAV. To provide meaningful data in such operations, all UAVs in a formation must use a single reference frame (typically the global frame). Traditionally, UAVs have access to the global frame via GPS. However, GPS signals may be lost in urban environments and enemy controlled airspace (jamming). 

Without access to global coordinates, a UAV must rely on its inertial navigation system (INS). Positions in the INS frame are updated by integrating measurements from gyroscopes and accelerometers. Error in these measurements causes the INS frame to accumulate drift. At any given time, drift is characterised by rotation and translation with respect to the global frame, and is assumed to be independent between UAVs in a formation. The error in gyroscope and accelerometer readings is stochastic. As a result, INS frame drift cannot be modelled deterministically. Information from both the global and INS frames must be collected and used to determine the drift between frames. This process is described as cooperative localisation when multiple vehicles provide information.

Various measurement types such as distance between agents and direction of arrival of a signal (we henceforth call DOA) can be used for this process. In the context of unmanned flight, additional sensors add weight and consume power. As a result, one generally aims to minimise the number of measurement types required for localisation. This paper studies a cooperative approach to localisation using DOA measurements.

%
%

We begin a review of existing literature by distinguishing localisation from tracking. A tracking problem typically involves updating a solution in real time, as new measurements are made available. Tracking is performed using Bayesian filters such as particle filters \citep{Lin2013}
and the Extended Kalman Filter \citep{Chang2014}.
In localisation, a set of measurements collected over a time interval is used to compute a solution at the end of the interval. A dynamic model is typically not required in localisation, and is not used in this paper. In \citep{Tekdas2010}, a target is localised using {\color{black} bearing}\footnote{The term bearing is used interchangeably with direction of arrival (DOA) in a two-dimensional ambient space. However, we refrain from using it in three-dimensional ambient space} measurements taken in the global frame by a stationary sensor network. In \citep{Duan2012}, self-localisation of a robot is achieved by taking {\color{black} bearing}-only measurements from the robot's INS frame towards a set of known global positions. \citep{Bayram2016} demonstrates how an agent is able to localise a stationary target using {\color{black} bearing}-only measurements. Optimal path planning algorithms allowing GPS-enabled sensing agents to localise a fixed target are studied in \citep{Lim2014}.
\citep{7815320} proposed a method of self-localisation of two agents using {\color{black} bearing} measurements but each agent was restricted to stationary two-dimensional circular orbit. Simultaneous localisation and mapping (SLAM) methods allow for self-localisation using only INS positions and {\color{black} bearing} measurements 
\citep{Bailey2003}. 
The local frame is anchored by positions of landmarks which are generally stationary in the global frame and known a priori or continuously sensed. 

The problem addressed in this paper is to localise a GPS-denied UAV, which we will call Agent B, with the assistance of a nearby GPS-enabled UAV, which we will call Agent A. Agent A broadcasts its global coordinates at discrete instants in time. Both agents move about arbitrarily in three-dimensional space. Using each broadcast of Agent A, Agent B is able to take a DOA measurement towards Agent A.

This problem and the solution proposed in this paper are novel. In particular, while the literature discussed above considers certain aspects from the following list, none consider all of the following aspects simultaneously:
\begin{itemize}
\item The network consists of only two mobile agents (and is therefore different to the sensor network localisation problems in the literature).
\item There is no a priori knowledge or sensing of a stationary reference point in the global frame.
\item The UAVs execute unconstrained arbitrary motion in three-dimensional space.
\item Cooperation between a GPS-enabled and a GPS-denied UAV (transmission of signals is unidirectional; from Agent A to Agent B.
\end{itemize}

\citep{7798924} studied this problem in two-dimensional space using bearing measurements. One single piece of data is acquired at each time step. As is detailed in the sequel, two scalar quantities are obtained with each DOA measurement in the three-dimensional case.

The rest of the paper is structured as follows. In Section~\ref{sec:problem_def} the problem is formalised. In Section~\ref{sec:noiseless} a solution to the noiseless case is presented. Section~\ref{sec:noisy_measurements} extends this method to allow for noise in DOA measurements. Section~\ref{sec:simulations} presents simulation results for the noisy case, studying the same example trajectory and a set of Monte Carlo simulations. The paper is concluded in Section~\ref{sec:conclusion}.

\section{Problem Definition}\label{sec:problem_def}

Two agents, which we call Agent A and Agent B, follow arbitrary trajectories in three-dimensional space. Agent A is able to self-localise in the global frame and therefore navigates with respect to the global frame. {\color{black}Because Agent B is GPS-denied, it has no self-localisation capacity in the global frame, but can self-localise in an INS frame. This INS frame drifts with respect to the global frame over long periods, but can be considered constant in short intervals.}

Positions each of agent {\color{black} in their navigation frames} and DOA measurements of signals between agents are obtained through a discrete-time measurement process. We let $P^I_J(k)$ denote the position of agent $J$ in the coordinate frame of agent $I$ at a $k^{th}$ time instant. Time between measurements can be variable. Let $u$, $v$, $w$ denote global frame coordinates, and $x$, $y$, $z$ denote coordinates in the INS frame of Agent B. Position coordinates at time instant $k$ are expressed as follows:
\begin{equation}
P_A^A(k) = [u_A(k) \; v_A(k) \; w_A(k)] ^\top 
\end{equation}
\begin{equation}
P_B^B(k) = [x_B(k) \; y_B(k) \; z_B(k)] ^\top
\end{equation}

Agent B's INS frame evolves via drift which, at a particular time instant, can be modelled as a rotation and a translation from the global frame. The rotation and origin of the INS frame is considered to be time invariant over short intervals, during which the following measurement process can occur multiple times. At each time instant $k$:
\begin{itemize}
\item Agent A records and broadcasts its position in the global frame, i.e. $P_A^A(k)$.
\item Agent B records its own position in the INS frame, i.e. $P_B^B(k)$.
\item Agent B receives the broadcast from Agent A, and measures the signal's DOA in the INS frame of B. Receipt of a broadcast may trigger the use of an optical sensor
\end{itemize}
We assume the delay between Agent A sending its coordinates and Agent B receiving this information is negligible, i.e. we assume the three activities occur simultaneously. 

The aim is to express the coordinates of Agent B in the global frame, in other words to determine:
\begin{equation}
P_B^A(k) = [u_B(k) \; v_B(k) \; w_B(k)] ^\top 
\end{equation}
Passing between the global frame and the INS frame of Agent B is achieved by a rotation of frame axes and translation of origin point of the frame. For instance, the coordinates vector of Agent A in the INS frame of Agent B is obtained as follows:
\begin{equation}
\label{eq:trans}
P_A^B = \bm{R^B_A} \, P_A^A + \bm{T_A^B}
\end{equation}

Therefore we have $P_B^A = \bm{R^{B\top}_A} \, (P_B^B - \bm{T_A^B})$ where $ \bm{R^{B\top}_A} = \bm{R^A_B} $ and $-\bm{R^{B\top}_A} \, \bm{T_A^B} = \bm{T_B^A}$. The localisation problem can therefore be reduced to solving for the entries of $\bm{R_A^B} \in SO(3)$ indexed $r_{ij}$ and $\bm{T_A^B} \in \mathbb{R}^3$ indexed $t_i$.
Direction of arrival measurements taken by Agent B towards Agent A are performed by instruments fixed to the body of Agent B. It is assumed that through the use of attitude measurements in the INS frame, Agent B is able to convert DOA readings from its body fixed frame to a body centred frame with the same orientation (rotation) as its INS frame.

Direction of arrival in Agent B's body centred frame with the same orientation as its INS frame is therefore expressed using the following spherical coordinate system:
\begin{itemize}
\item Azimuth ($\theta$): angle formed between the positive $x$ axis and the projection the vector from Agent B towards Agent A onto $xy$ plane.
\item Elevation ($\phi$): angle formed between unit vector towards Agent A and $xy$ plane. The angle is positive if the $z$ component of the unit vector towards Agent A is positive.
\end{itemize}

The total number of scalar measurements required to solve the localisation problem cannot be smaller than the number of degrees of freedom in the rotation matrix and translation vector. Note that $\bm{R_A^B}$ is a rotation matrix ($\bm{R_A^B} \in SO(3)$), with possible parametrisations including Euler angles and Rodrigues parametrisation. We work directly with entries $r_{ij}$ to obtain linear problems. Euler angle parametrisation of $\bm{R_A^B}$ is a 3 vector. The translation vector is a 3 vector. This problem therefore has 6 degrees of freedom.

The matrix $\bm{R_A^B}$ is a rotation matrix if and only if $\bm{R_A^B}\bm{R_A^{B\top}} = I_3$ and $\det(\bm{R_A^B}) = 1$. As will be seen in the sequel, these constraints are equivalent to a set of quadratic constraints on entries of $\bm{R_A^B}$. In total there are 12 entries of $\bm{R^B_A}$ and $\bm{T^B_A}$ to be found. An equation set of $n$ independent relations (including measurements and constraints) for $n$ variables will generically have multiple solutions if at least one of the relations is quadratic. When an additional scalar measurement is taken, generically a unique solution exists.


\section{Localisation with Noiseless Measurements}\label{sec:noiseless}

This section considers noiseless position and direction of arrival measurements. Section 4 considers noise in direction of arrival measurements only. The consideration of noisy position measurements is left for future work.

\subsection{Forming a system of linear equations}

A unit vector in the INS frame, pointing from Agent B to Agent A, defined by azimuth and elevation angles $\theta$ and $\phi$, can be parametrised as follows:
\begin{equation}
\hat{d}(\theta, \phi)=
\begin{bmatrix}
    \cos\theta\cos\phi, &
    \sin\theta\cos\phi, 
    &
    \sin\phi 
\end{bmatrix}^\top
\end{equation}
The following static analysis holds for all instants in time, hence the specification $k$ is dropped. Define $\bar d \doteq \Vert P_B^A - P_B^B\Vert$ as the Euclidean distance between Agent A and Agent B (which is not available to either agent). Scaling the unit vector $\bar d$ gives
\begin{equation}
\label{eq:system1}
\hat{d}(\theta, \phi)
=
\frac{1}{\bar d}
\begin{bmatrix}
    x_A - x_B, &
    y_A - y_B, &
    z_A - z_B &
\end{bmatrix}^\top
\end{equation}
Applying equation (\ref{eq:trans}) yields:
\begin{equation}
\label{eq:scaled}
\begin{bmatrix}
    \cos\theta
    \cos\phi \\
    \sin\theta\cos\phi \\
    \sin\phi \\
\end{bmatrix}
=
\frac{1}{\bar d}
\begin{bmatrix}
    r_{11} u_A + r_{12} v_A + r_{13} w_A + t_1 - x_B \\
    r_{21} u_A + r_{22} v_A + r_{23} w_A + t_2 - y_B \\
    r_{31} u_A + r_{32} v_A + r_{33} w_A + t_3 - z_B \\
\end{bmatrix}
\end{equation}
The left hand vector is calculated directly from DOA measurements. Each entry of the vector on the right hand side is a linear combination of entries of $\bm{R^B_A}$ and $\bm{T^B_A}$. Cross-multiplying entries 1 and 3 of both vectors eliminates $\bar d$, and yields the following equality:
\begin{align}
\begin{split}
\label{xzlinka}
\sin{\phi}(r_{11} u_A + r_{12} v_A + r_{13} w_A + t_1 - x_B) = \\
\cos{\theta}\cos{\phi}(r_{31} u_A + r_{32} v_A + r_{33} w_A + t_3 - z_B) 
\end{split}
\end{align}
which for convenience is rearranged as
\begin{align}
\begin{split}
\label{eq:xzlinkb}
& (u_A\sin{\phi})r_{11} + (v_A\sin{\phi})r_{12} + (w_A\sin{\phi})r_{13} \\
& - (u_A\cos{\theta}\cos{\phi})r_{31} - (v_A\cos{\theta}\cos{\phi})r_{32} \\
& - (w_A\cos{\theta}\cos{\phi})r_{33} + (\sin{\phi})t_1 - (\cos{\theta}\cos{\phi})t_3 \\
& = (\sin{\phi})x_B - (\cos{\theta}\cos{\phi})z_B
\end{split}
\end{align}
Similarly, cross-multiplying entries 2 and 3 of both vectors appearing in equation (\ref{eq:scaled}) yields a similar equation.
\begin{align}
\begin{split}
\label{eq:yzlinkb}
& (u_A\sin{\phi})r_{21} + (v_A\sin{\phi})r_{22} + (w_A\sin{\phi})r_{23} \\
& - (u_A\sin{\theta}\cos{\phi})r_{31} - (v_A\sin{\theta}\cos{\phi})r_{32} \\
& - (w_A\sin{\theta}\cos{\phi})r_{33} + (\sin{\phi})t_2 - (\sin{\theta}\cos{\phi})t_3 \\
& = (\sin{\phi})y_B - (\sin{\theta}\cos{\phi})z_B
\end{split}
\end{align}
Both equations 
\eqref{eq:xzlinkb} and \eqref{eq:yzlinkb} 
are linear in the unknown $r_{ij}$ and $t_i$ terms. Given a series of $K$ measurements, these equations can be used to construct the following system of linear equations:
\begin{equation}
\label{eq:Axb}
A\Psi = b
\end{equation}
where the 12-vector of unknowns $\Psi$ is defined as:
\begin{equation}
\Psi = [r_{11} \enspace r_{12} \enspace r_{13} \enspace ... \enspace r_{31} \enspace r_{32} \enspace r_{33} \enspace t_1 \enspace t_2\enspace t_3]^\top
\end{equation}
The matrix $A$ contains the known values $P_A^A(k)$ and $\phi(k), \theta(k)$ for $k = 1, ..., K$. The vector $b$ contains the known values $P_B^B(k)$ and $\phi(k), \theta(k)$ for $k = 1, ..., K$. The precise forms of $A$ and $b$ are omitted due to spatial limitations, and will be included in an extended version of the paper.


\subsection{Example with noiseless DOA measurements}\label{ssec:example_noiseless}
 
If $K \geq 6$, the matrix $A$ will be square or tall. In the noiseless case, if $A$ is of full column rank, equation (\ref{eq:Axb}) will be solvable.
We demonstrate this using a set of example trajectories detailed in Table~\ref{tab:noiseless_data}. We will make additional use of this example trajectory in the noisy measurement case presented in Section~\ref{sec:simulations}. Extensive Monte Carlo simulations demonstrating localisation for a large number of generic flight trajectories are left to the noisy measurement case. Nongeneric flight trajectories are discussed immediately below, in subsection~\ref{ssec:nongeneric_traj}.

Example trajectories for Agents A and B in the global frame were generated. These trajectories satisfied a set of flight properties to be included in the extended version of this paper. These are plotted in Figure \ref{fig:example_trajectory}. The rotation matrix and translation vector used in simulation are:
\begin{equation}
\bm{R_A^B} = 
\begin{bmatrix}
    -0.627 & -0.776 & 0.072 \\
    -0.747 & 0.625 & 0.228 \\
    -0.222 & 0.090 & -0.971
\end{bmatrix}
\end{equation}
\begin{equation} 
\bm{T_A^B} = 
\begin{bmatrix}
	247.490 \quad 110.382 \quad 229.784 \\
\end{bmatrix} ^\top
\end{equation}

Table \ref{tab:noiseless_data} includes azimuth and elevation angle measurements from Agent B towards Agent A in the measurement frame, for $K = 6$ time instants. Using \eqref{eq:Axb}, $\bm{R_A^B}$ and $\bm{T_A^B}$ were solved exactly for the given flight trajectories.

\begin{table}
\caption{Positions of Agents A and B in their respective frames and noiseless DOA measurements for example trajectories}
\label{tab:noiseless_data}
\resizebox{\linewidth}{!}{%
\begin{tabular}{||c c c||} 
 \hline
 \vspace{0.5mm}
 Time $k$ & $P_A^A$ & DOA [$\theta$ $\phi$] \\
 & $P_B^B$ & \\[0.5ex] 
 \hline\hline
 1 & [$0$ \: $0$ \: $300$]$^\top$ & [-1.500 -0.851] \\
 & [$89.680$ \: $1035.199$ \: $474.865$]$^\top$ & \\
 \hline
 2 & [$82.962$ \: $-235.407$ \: $314.161$]$^\top$ & [-1.679 -0.835] \\
 & [$-40.633$ \: $1157.514$ \: $649.672$]$^\top$ & \\
 \hline
 3 & [$141.084$ \: $-478.270$ \: $302.352$]$^\top$ & [-1.789 -0.812] \\
 & [$-182.218$ \: $1255.810$ \: $830.757$]$^\top$ & \\
 \hline
 4 & [$139.079$ \: $-726.308$ \: $271.157$]$^\top$ & [-1.977 -0.733] \\
 & [$-165.197$ \: $1416.859$ \: $1021.213$]$^\top$ & \\
 \hline
 5 & [$-109.876$ \: $-704.457$ \: $277.792$]$^\top$ & [-1.828 -0.790] \\
 & [$-217.778$ \: $1581.963$ \: $1201.424$]$^\top$ & \\
 \hline
 6 & [$-252.217$ \: $-499.403$ \: $291.634$]$^\top$ & [-1.499 -0.833] \\
 & [$-452.605$ \: $1649.158$ \: $1254.728$]$^\top$ & \\
 \hline
\end{tabular}}
\end{table}
\vspace{2mm}

\subsection{Nongeneric trajectories}\label{ssec:nongeneric_traj}
It is obvious that we can only localise Agent B by solving \eqref{eq:Axb} if matrix $A$ has full column rank. We are therefore motivated to identify trajectories of Agent A and Agent B which result in $\text{rank}(A) < 6$ (assuming $K \geq 6$) so we can avoid them. A large number of simulations revealed that $\text{rank}(A) < 6$ is consistently encountered for all $K \geq 6$ when the trajectory of Agent A is restricted to a plane. Note that this plane need not be parallel to a coordinate plane of either the global or INS coordinate frames. We leave further, detailed analytical treatment of nongeneric trajectories to the extended version of this paper.
\section{Localisation with Noisy DOA Measurements}\label{sec:noisy_measurements}

In practice, DOA measurements from sensors on Agent B will be contaminated with noise. This section develops an approach to deal with the noisy measurements which are encountered in real-world implementation. We approach the problem using semidefinite programming to exploit the quadratic constraints on $\Psi$ arising from the properties of the rotation matrix. Such constraints would not be utilised in a standard unconstrained least squares approach.

In practice, noise contaminated azimuth and elevation measurements are taken in the body fixed frame of Agent B. Strictly speaking, the noise is expected to follow a von Mises distribution \citep{forbes2011statistical_book}. For small noise, such as what we encounter, the von Mises distribution can be approximated by a Gaussian distribution. The noise affecting azimuth and elevation are assumed to be independent when measured in the body fixed frame, however they are not known to have equal standard deviations. These noises are likely to lose independence when azimuth and elevation measurements measured in the body fixed frame are converted into azimuth and elevation values in Agent B's body centred frame with the same orientation as its INS frame.

\subsection{Quadratic constraints on entries of $\Psi$}\label{ssec:constraints_psi}

We now identify 21 quadratic constraints on entries of $\bm{R_A^B}$. Note that this set is not independent, as discussed in Remark~\ref{rem:dependent_constraints} below. Recall the orthogonality property $\bm{R_A^B}\bm{R_A^{B\top}} = I_3$. By computing each entry of $\bm{R_A^B}\bm{R_A^{B\top}}$ and setting these equal to entries of $I_3$, we define constraints $C_i$ for $i = 1, ..., 6$:
\begin{subequations}\label{eq:constraints_psi_01}
\begin{align}
& C_1 = \psi_1^2 + \psi_2^2 + \psi_3^2 - 1 = 0 \\
& C_2 = \psi_4^2 + \psi_5^2 + \psi_6^2 - 1 = 0 \\
& C_3 = \psi_7^2 + \psi_8^2 + \psi_9^2 - 1 = 0 \\
& C_4 = \psi_1\psi_4 + \psi_2\psi_5 + \psi_3\psi_6 = 0 \\
& C_5 = \psi_1\psi_7 + \psi_2\psi_8 + \psi_3\psi_9 = 0 \\
& C_6 = \psi_4\psi_7 + \psi_5\psi_8 + \psi_6\psi_9 = 0
\end{align}
\end{subequations}
To simplify notation we call $C_{j:k}$ the set of constraints $C_i$ for $i = j, .., k$. Similarly, by computing each entry of $\bm{R_A^{B\top}}\bm{R_A^{B}}$ and setting these equal to $I_3$, we define constraints $C_{7:12}$. We omit presentation of $C_{7:12}$ due to spatial limitations and similarity with $C_{1:6}$. The sets $C_{1:6}$ and $C_{7:12}$ are clearly equivalent.

Further constraints are required to ensure $\det(\bm{R_A^B}) = 1$. Cramer's formula states that
${\bm{R^B_A}}^{-1} = \text{adj}(\bm{R^B_A})/\det(\bm{R_A^{B}})$,
where $\text{adj}(\bm{R^B_A})$ denotes the adjugate matrix of $\bm{R_A^{B}}$. Orthogonality of $\bm{R_A^{B}}$ implies ${\bm{R^B_A}}^\top = \text{adj}(\bm{R^B_A})$ or that $\bm{R^B_A} = \text{adj}(\bm{R^B_A})^\top$. By computing entries of the first column of $Z = \bm{R_A^{B}} - \text{adj}(\bm{R^B_A})^\top$ and setting these equal to 0, we define constraints $C_{13:15}$:
\begin{subequations}\label{eq:constraints_psi_02}
\begin{align}
& C_{13} = \psi_1 - (\psi_5\psi_9 - \psi_6\psi_8) = 0 \\
& C_{14} = \psi_4 - (\psi_3\psi_8 - \psi_2\psi_9) = 0 \\
& C_{15} = \psi_7 - (\psi_2\psi_6 - \psi_3\psi_5) = 0
\end{align}
\end{subequations}
Similarly, by computing the entries of the second and third columns of $Z$ and setting these equal to 0, we define constraints $C_{16:18}$ and $C_{19:21}$. Due to space limitations, we omit presenting them. The complete set $C_{1:21}$ constrains $\bm{R_A^{B}}$ to be a rotation matrix. We refer to this set as $C_{\Psi}$.

Due to these additional relations, localisation requires azimuth and elevation measurements at 4 instants only ($K = 4$), as opposed to 6 instants required in Section \ref{sec:noiseless}.

\subsection{Formulation of the Semidefinite Program}\label{ssec:sdp_def}
The goal of the semidefinite program is to obtain:
\begin{equation}
\label{eq:non-sdp_objective}
\argmin_\Psi ||A\Psi-b||
\end{equation}
subject to $C_{\Psi}$. Equivalently, we seek $\argmin_\Psi ||A\Psi-b||^2$ subject to to $C_{\Psi}$. We define the inner product of two matrices $U$ and $V$ as $\langle U,V \rangle = \text{trace}(U,V^\top)$. One obtains
\begin{align}
||A\Psi-b||^2 = \left\langle P , X \right\rangle
\end{align}
where $ P = 
\begin{bmatrix}
    A \quad b
\end{bmatrix}^\top
\begin{bmatrix}
    A \quad b
\end{bmatrix}
$ and 
\begin{equation}
\label{eq:def_of_X}
X = 
\begin{bmatrix}
    \Psi \\
    -1 \\
\end{bmatrix}
\begin{bmatrix}
    \Psi \\
    -1 \\
\end{bmatrix} ^\top
\end{equation}
and $X$ is a rank 1 positive-semidefinite matrix\footnote{All matrices $M$ which can be expressed in the form of $M = vv'$ where $v$ is a column vector are positive-semidefinite matrices.}. The constraints $C_{\Psi}$ can also be expressed in inner product form. For $i = 1, ..., 21$, $C_i = 0$ is equivalent to $\langle Q_i, X \rangle = 0$ for some $Q_i$. Solving for $\Psi$ in (\ref{eq:non-sdp_objective}) is therefore equivalent to solving for:
\begin{equation}
\argmin_\Psi \langle P, X \rangle
\end{equation}
such that:
\begin{equation}
\label{eq:sdpconstr1}
X \geq 0
\end{equation}
\begin{equation}
\label{eq:sdpconstr2}
\text{rank}(X) = 1
\end{equation}
\begin{equation}
\label{eq:sdpconstr3}
X_{13,13} = 1
\end{equation}
\begin{equation}
\langle Q_i, X \rangle = 0
\end{equation}
for $i = 1, ..., 21$.
This fully defines the semidefinite program.

\begin{rem}[Inclusion of Dependent Constraints]\label{rem:dependent_constraints}\hfill 
Note that while the set of equations \eqref{eq:constraints_psi_01} and \eqref{eq:constraints_psi_02} form an independent set of constraints on the first nine entries of $\Psi$, the set of constraints $C_{\Psi}$ is not independent.  For instance,
$C_{1:6}$ and $C_{7:12}$ are equivalent. For simplicity, we use $C_{\Psi}$ in the SDP formulation in this paper. In the extended version of this paper, we will thoroughly investigate whether localisation accuracy improves with the additional dependent constraints.
\end{rem}

\begin{rem} [Normalisation of translation terms]
It was observed that SDP solution accuracy deteriorates when the condition number of $X$ of the true solution is high. When approximate size of $\bm{T_A^B}$ is known, coefficients associated to entries $t_i$ in equations (\ref{eq:xzlinkb}) and (\ref{eq:yzlinkb}) are scaled to solve for values close to 1.
\end{rem}

\subsection{Rank Relaxation of Semidefinite Program}\label{ssec:relax_sdp}

This semidefinite program is a reformulation of a quadratically constrained quadratic program (QCQP). Computationally speaking, QCQP problems are generally NP-hard. A close approximation to the true solution can be obtained in polynomial time if the rank 1 constraint on $X$ is relaxed. The solution to the relaxed semidefinite program $X$ is typically close to being a rank 1 matrix\footnote{The measure used for closeness to rank 1 is the ratio of the two largest singular values in the singular value decomposition of $X$.}. The closest rank 1 approximation to $X$, which we call $\hat{X}$, is obtained by evaluating the singular value decomposition of $X$, then setting all singular values except the largest equal to zero.

From $\hat{X}$, one can then use the definition of $X$ in (\ref{eq:def_of_X}) to obtain the approximation of $\Psi$, which we will call $\hat\Psi$. Entries $\hat\psi_{i}$ for $i = 10, 11, 12$ can be used immediately to construct an estimate for $\bm{T_A^B}$, which we will call $\overline{T}$. Entries $\hat\psi_{i}$ for $i = 1, ..., 9$ will be used to construct an approximation of $\bm{R_A^B}$, which we will call $\widehat{R}$.

\subsection{Orthogonal Procrustes Problem}\label{ssec:procrustes}
Due to the relaxation of the rank constraint on $X$, it is no longer guaranteed that entries of $\hat\Psi$ strictly satisfy the set of constraints $C_{\Psi}$. Specifically, the matrix $\widehat{R}$ may not be a rotation matrix. The Orthogonal Procrustes algorithm is a commonly used tool to determine the closest orthogonal matrix (denoted $\overline{R})$ to a given matrix, $\widehat{R}$. This is given by $\overline{R} = \argmin_\Omega ||\Omega - \widehat{R}||_F$ subject to $\Omega\Omega^\top = I$, where $||.||_F$ is the Frobenius norm. 

When noise is high, the above method occasionally returns $\overline{R}$ such that $\det(\overline{R}) = -1$. A special case of the Orthogonal Procrustes algorithm is available to ensure that $\det(\overline{R}) = 1$, i.e. $\overline{R}\in SO(3)$ is a proper rotation matrix but we omit the minor details due to spatial limitations.

The matrix $\overline{R}$ and vector $\overline{T}$ are the final approximations of $\bm{R_A^B}$ and $\bm{T_A^B}$ using the combination of the semidefinite program and Orthogonal Procrustes methods. The estimate of the position of B in the global frame is $\overline{P_B^A} = \overline{R}^\top(P_B^B - \overline{T})$.

\subsection{Metrics for error in $\overline{R}$ and $\overline{T}$}\label{ssec:error_metrics}

This paper uses the geodesic metric for rotation.
This metric on $SO(3)$ defined by
\begin{equation}
d(R_1,R_2) = \arccos\bigg(\frac{\text{tr}(R_1^\top R_2)-1}{2}\bigg)
\end{equation}
is the magnitude of angle of rotation about this axis \citep{7515271}. Where $\bm{R_A^B}$ is known, the error of rotation $\overline{R}$ is defined as $d(\overline{R},\bm{R_A^B})$. 
Position error is defined as the average Euclidian distance between true global coordinates of B, and estimated global coordinates over the $K$ measurements taken.
\begin{equation}
error(\overline{P_B^A}) = \frac{\sum_k ||\bm{R^{B\top}_A} \, (P_B^B(k) - \bm{T_A^B}) - P_B^A(k)||}{K}
\end{equation}

\section{Simulation Results}\label{sec:simulations}

In this section, we investigate the performance of the SDP algorithm combined with the Orthogonal Procrustes Algorithm in the presence of noise (we shall call this combined algorithm SDP+O for short). We acknowledged in Section~\ref{sec:noisy_measurements} that noise in measurements of $\theta(k)$ and $\phi(k)$ may not have the same standard deviations. However, for simplicity and with space limitations in mind, we will assume in this paper that the standard deviations are the same (which we denote as $\sigma$). In other words, $\widetilde{\theta}(k) = \theta(k) + \zeta_1$ and $\widetilde{\phi}(k) = \phi(k) + \zeta_2$ where $\zeta_1, \zeta_2 \sim N(0, \sigma^2) $ Extensive exploration of different standard deviations for $\theta(k)$ and $\phi(k)$ will be conducted in this paper's extended version.

\subsection{Example with noisy DOA measurements}\label{ssec:example_noisy}

\begin{figure}
\centering
\includegraphics[width=0.9\linewidth]{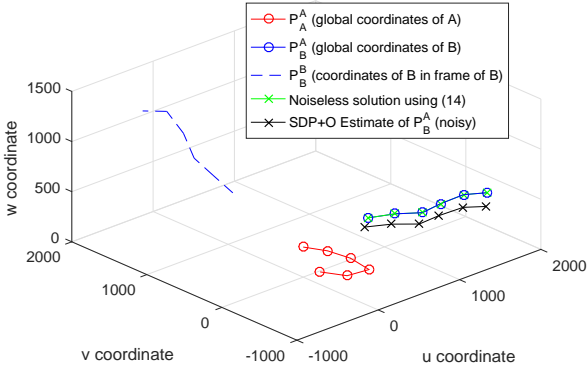}
\caption{Recovery of global coordinates of Agent B for example trajectories (error of $\sigma = 3^{\circ}$ in noisy case)}
\label{fig:example_trajectory}
\end{figure}

Samples of Gaussian error with $\mu = 0^\circ$, $\sigma = 3^\circ$ were added to elevation and azimuth measurements simulated in Section 3.2. The SDP+O algorithm was used to obtain of $\overline{R}$ and $\overline{T}$. The reconstructed trajectory $\overline{P_B^A}$ is plotted in Figure \ref{fig:example_trajectory}. Position data of the reconstructed trajectory $\overline{P_B^A}$ and simulated noise contaminating azimuth and elevation readings is tabulated in Table \ref{tab:noisy_data}.

\begin{table}
\caption{Positions of $\overline{P_B^A}$ and ${P_B^A}$, and Gaussian noise ($\sigma = 3^{\circ}$) added to azimuth and elevation measurements ($\zeta_1, \zeta_2 \sim N(0, \sigma^2)$)}
\label{tab:noisy_data}
\resizebox{\linewidth}{!}{%
\begin{tabular}{||c c c||} 
 \hline
 \vspace{0.5mm}
 Time $k$ & $P_B^A$ & [$\zeta_1 \: \zeta_2$] [rads] \\
 & $\overline{P_B^A}$ & [$\zeta_1 \: \zeta_2$] [degs]\\[0.5ex] 
 \hline\hline
 1 & [$800$ \: $0$ \: $350$]$^\top$ & [0.0747 -0.0637] \\
 & [$695.7$ \: $-61.6$ \: $309.1$]$^\top$ & [4.2809 -3.6483]\\
 \hline
 2 & [$1017.5$ \: $-122.5$ \: $364.1$]$^\top$ & [-0.0350 0.0699] \\
 & [$902.6$ \: $-201.4$ \: $321.9$]$^\top$ & [-2.0029 4.0065]\\
 \hline
 3 & [$1225.7$ \: $-260.8$ \: $358.0$]$^\top$ & [-0.0478 -0.1032] \\
 & [$1098.9$ \: $-355.9$ \: $314.3$]$^\top$ & [-2.7416 -5.9145]\\
 \hline
 4 & [$1474.1$ \: $-233.5$ \: $363.9$]$^\top$ & [0.0243 0.0084] \\
 & [$1348.7$ \: $-348.8$ \: $320.9$]$^\top$ & [1.3943 0.4801]\\
 \hline
 5 & [$1719.3$ \: $-272.8$ \: $392.6$]$^\top$ & [-0.0247 0.0165] \\
 & [$1589.9$ \: $-408.2$ \: $349.5$]$^\top$ & [-1.4124 0.9474]\\
 \hline
 6 & [$1810.6$ \: $-496.1$ \: $458.0$]$^\top$ & [0.0937 -0.0031] \\
 & [$1662.8$ \: $-639.0$ \: $412.0$]$^\top$ & [5.3698 -0.1750]\\
 \hline
\end{tabular}}
\end{table}
\vspace{2mm}


\subsection{Monte Carlo Simulations on Random Trajectories}\label{ssec:rand_traj}

In this subsection, we conduct extensive Monte Carlo simulations to show the viability of the SDP+O algorithm. We randomly generated 100 \emph{realistic fixed wing UAV trajectories}. By realistic, we mean that the distance separation between successive measurements is consistent with UAV flight speeds and ensures the UAV does not exceed an upper bound on the turn/climb rate. We vary the noise level by $\sigma = 0.1^\circ, 1^\circ, 2^\circ, 3^\circ, 5^\circ$. The noise level $\sigma = 0.1^\circ$ is representative of an optical sensor, while other noise levels are representative of antenna-based (RF) measurements. For each level of noise, we conducted 100 Monte Carlo simulations with unique trajectories. For each noise level, we then calculated the median $d(\overline{R},{\bm{R^B_A}})$ and $error(\overline{P_B^A})$ over the 100 simulations.



\begin{figure}
\includegraphics[width=\linewidth]{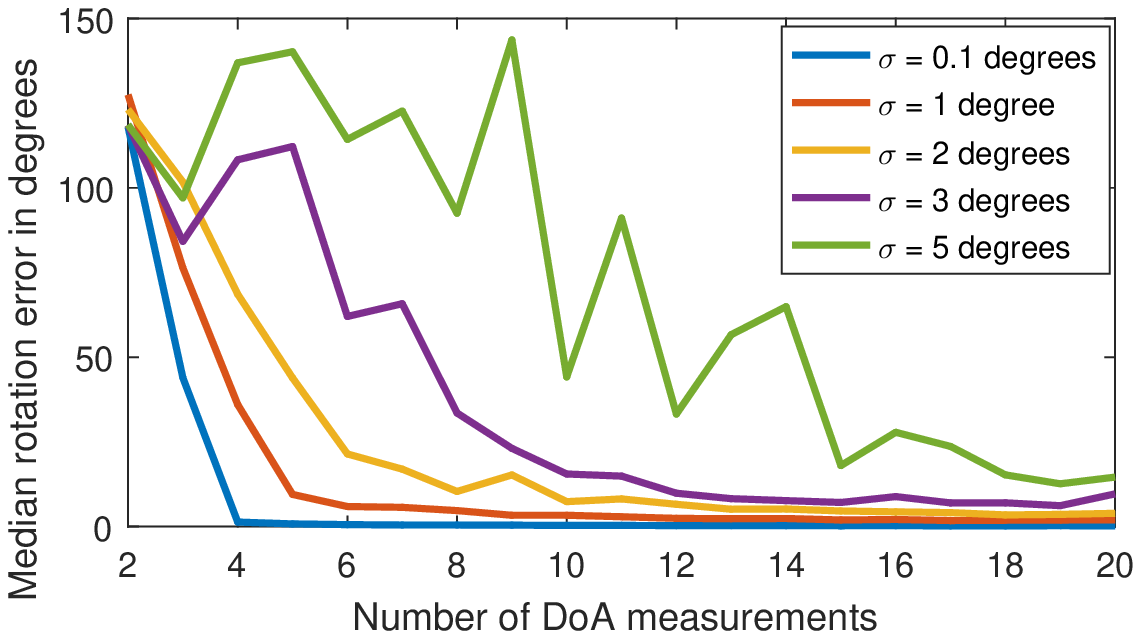}
\caption{Median $d(\overline{R},{\bm{R^B_A}})$ vs. number of DOA measurements used to solve SDP+O. }
\label{fig:rotation_ks}
\end{figure}

\begin{figure}
\includegraphics[width=\linewidth]{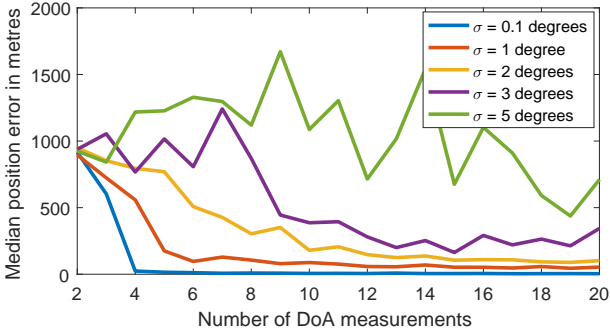}
\caption{Median $error(\overline{P_B^A})$ vs. number of DOA measurements used to solve SDP+O (average distance of 1.38km between agents).}
\label{fig:translation_ks}
\end{figure}



Median is a commonly used as measure of central tendency to mitigate the effect of outliers. We use this measure to analyse Monte Carlo simulation results.
Discussion on the cumulative frequency distribution of errors obtained for each set of 100 Monte Carlo simulations will be included in the extended version of this paper.

The median $d(\overline{R},{\bm{R^B_A}})$ and $error(\overline{P_B^A})$ versus the number of measurements (i.e. $K$) used to solve the SDP+O are plotted in Figure~\ref{fig:rotation_ks} and Figure~\ref{fig:translation_ks}, respectively. It is observed that for each value of $\sigma$ there is greater improvement in rotation error than translation error as $K$ is increased. This will be discussed further in the extended version of this paper.

\subsection{Comparison of LS+O and SDP+O}\label{ssec:lso_vs_sdpo}


In this subsection, we compare the SDP+O algorithm with an unconstrained Least Squares algorithm combined with the Orthogonal Procrustes algorithm (which we call LS+O). By LS+O, we mean that $\Psi^* = \argmin_\Psi \Vert A\Psi - b \Vert$ is found without any constraints on $\Psi$. The first nine entries of $\Psi$ are then used to form a matrix, and the Orthogonal Procrustes algorithm detailed in subsection~\ref{ssec:procrustes} is applied to obtain the closest rotation matrix $\overline{R}$.

\begin{figure}
\centering
\includegraphics[width=\linewidth]{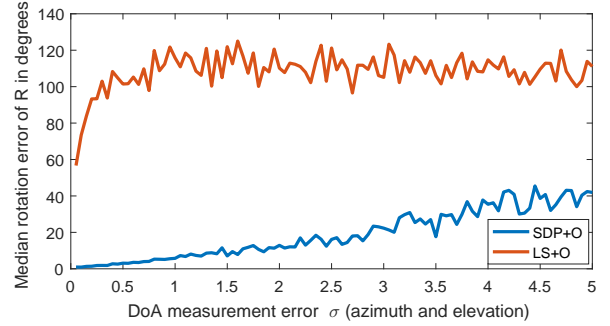}
\caption{Comparison of median $d(\overline{R},\bm{R_A^B})$ using SDP+O and LS+O methods for example trajectory ($K$ = 6).}
\label{fig:LSEvsSDP}
\end{figure}

Agents A and B are prescribed the representative trajectories illustrated in Figure 1. Noise level was varied from $\sigma = 0^\circ$ to $5^\circ$. For each noise level, we conducted 20 Monte Carlo simulations of the same trajectory using LS+O and SDP+O. The error index $d(\overline{R},{\bm{R^B_A}})$ was separately calculated over the 20 simulations for LS+O and SDP+O. Figure~\ref{fig:LSEvsSDP} shows the median $d(\overline{R},{\bm{R^B_A}})$ versus level of noise for both LS+O and SDP+O. The median $d(\overline{R},{\bm{R^B_A}})$ using the LS+O algorithm was consistently double that using the SDP+O method. Median errors in the order of $90^\circ$ are expected from a random set of rotation matrices.
We conclude that SDP+O is the superior method. 
Its median performance is approximately linear in the mean azimuth and elevation measurement error standard deviation.

\section{Conclusion}\label{sec:conclusion}
This paper studied a cooperative localisation problem between a GPS-denied and a GPS-enabled UAV. The GPS-enabled UAV broadcasts it position in discrete time to the GPS-denied UAV, and the GPS-denied UAV measures the direction of arrival of the broadcast signal. It was found that localisation could be achieved by solving a non-linear system with less than 6 DOA measurements. A localisation algorithm was developed in two stages. With noiseless DOA measurements, we showed that a linear system of equations built from six or more measurements yielded the localisation solution for generic trajectories. The second stage considered noisy DOA measurements. An optimisation problem was defined in which the objective function and constraints on its arguments were quadratic. This problem therefore lent itself to quadratically constrained quadratic programming, which was solved numerically using a rank relaxed semidefinite program with the solution adjusted using the Orthogonal Procrustes algorithm. Simulations were presented to illustrate the algorithm when DOA measurements were noisy. Future work includes improving the reliability of the SDP+O method, studying multiple agents in a formation and filtering methods.


\bibliography{Literature2}

\end{document}